\documentclass[10pt, twocolumn]{IEEEtran}
\usepackage[T1]{fontenc}
\usepackage{color}
\usepackage{float}
\usepackage{mathrsfs}
\usepackage{amsmath}
\usepackage{amssymb}
\usepackage{graphicx}
\usepackage{booktabs} % 用于美化表格
\usepackage{caption}  % 控制表格标题格式
\usepackage{CJK}
\usepackage{indentfirst}
\usepackage{amsmath}
\usepackage{cases}
\usepackage{setspace}
\usepackage[unicode=true,
bookmarks=true,bookmarksnumbered=true,
bookmarksopen=true,bookmarksopenlevel=1,
breaklinks=false,pdfborder={0 0 0},
pdfborderstyle={},backref=false,
colorlinks=false]
{hyperref}
\hypersetup{pdftitle={Your Title},
pdfauthor={Yulu Han},
pdfpagelayout=OneColumn, 
pdfnewwindow=true, 
pdfstartview=XYZ, 
plainpages=false}
\makeatletter
\floatstyle{ruled}
\newfloat{algorithm}{tbp}{loa}
\providecommand{\algorithmname}{Algorithm}
\floatname{algorithm}{\protect\algorithmname}
\usepackage{setspace}
\usepackage{multirow} 
\usepackage{xcolor}
\usepackage[caption=false,font=footnotesize]{subfig}
\usepackage{algorithm}
\usepackage{algorithmic}
\usepackage{multirow} %multirow for format of table 
\usepackage{amsmath} 
\usepackage{xcolor}

\usepackage[caption=false,font=footnotesize]{subfig}
\allowdisplaybreaks[4]
\usepackage{cite}
\usepackage{bm}
\usepackage{makecell}
\usepackage{fancyhdr}
\usepackage{diagbox}
\usepackage{algorithmic}
\usepackage{algorithm}
\usepackage{graphicx}
\usepackage{enumitem}
\IEEEoverridecommandlockouts

\usepackage{lettrine}

\usepackage{geometry}
\usepackage{subfig}
\usepackage{hyperref}
\makeatother

\begin{document}
% \newgeometry{a4paper, left=1.33cm, right=1.33cm, bottom=4.31cm, top=1.95cm}
% \newgeometry{left=1in, right=1in, bottom=1in, top=1in}
\title{\textcolor{black}{CKM-Driven Communication-Aware UAV Intelligent Trajectory Optimization for Urban Inspection}}%CKM-Driven UAV Path Planning: Balancing distance and communication
\author{
\IEEEauthorblockN{Xiaomeng Yang$^{\dagger}$, Ziye Jia$^{\dagger}$, Qiuming Zhu$^{\dagger}$ and Qihui Wu$^{\dagger}$\\
}
\IEEEauthorblockA{
\small$^{\dagger}$The Key Laboratory of Dynamic Cognitive System of Electromagnetic Spectrum Space, Ministry of Industry and\\ 
\small Information Technology, Nanjing University of Aeronautics and  Astronautics, Nanjing, Jiangsu, 211106, China\\
\{yangxiaomeng, jiaziye, zhuqiuming, wuqihui\}@nuaa.edu.cn.
}
\thanks{This work was supported by National Key R\&D Program of China 2025YFF0514704. (Corresponding author: Ziye Jia)}
}

\maketitle
\thispagestyle{empty}
\begin{abstract}
Unmanned aerial vehicles (UAVs) are increasingly employed in urban inspection tasks, where reliable communication 
is critical but challenging due to the severe spatial channel heterogeneity. 
To address the issue, in this paper, we focus on the communication-aware path planning for multi-UAV tasks, and propose a channel knowledge map (CKM)-driven trajectory 
planning framework which integrates the channel modeling and trajectory decision-making. 
Specifically, we apply the diffusion model to construct a time-accumulated CKM and achieve the accurate 
perception with low flight overhead, which leverages the sparse observation data to reconstruct the high-fidelity global channel 
quality distribution.
Based on the CKM, we propose a global-to-local graph attention network soft actor-critic algorithm.
The graph attention network optimizes the complex combinatorial node ordering problem, generating an optimal and communication-aware sequence for the inspection targets.
Subsequently, the soft actor-critic algorithm performs continuous action control to ensure the smoothness of the flight path and dynamically avoid communication attenuation areas.
Simulation results demonstrate that the proposed method effectively guides UAVs through high-quality channel regions without dependence on real-time channel feedback, 
significantly improving both the trajectory efficiency and communication reliability.

\end{abstract}

\section{Introduction}
\vspace{2mm}
\lettrine[lines=2]
THE unmanned aerial vehicle (UAV) has emerged as a flexible platform showing unique advantages in urban 
inspection tasks \cite{11450383,11457394}. %,javed2023uav,9373545, including traffic monitoring, aerial data collection, etc
% The high mobility of UAVs enables adaptive deployment and flexible area coverage, showing unique advantages in urban inspection tasks. 
However, ensuring reliable air-to-ground communication during the task remains challenging, particularly in urban environments where 
dense buildings, street canyons, and dynamic blockages may cause severe path loss and rapid channel 
variations\cite{yang2023uav}. %,luo2023multi,ma2022automated
Thus, it is crucial to perform communication-aware path planning for urban inspection tasks.

The conventional UAV path planning approaches typically optimize geometric metrics such as distance or energy consumption, 
while neglecting essential communication constraints or assuming simplified channel models\cite{11159633}. %10759651,fan2023uav,
These assumptions fail to capture the complex spatial heterogeneity of wireless propagation, potentially leading 
UAVs to traverse regions with severely degraded communication quality.
% \enlargethispage{-\baselineskip}
% \enlargethispage{-\baselineskip}
To address this issue, increasing attention has been paid to the communication-
aware UAV path planning, where 
the knowledge of the radio environment is incorporated into trajectory design. 
For instance, \cite{zhang2020radio} preconstructs a signal to interference plus noise ratio (SINR) map
to determine flyable space meeting a given SINR threshold.
% For instance, in \cite{zhang2020radio}, a radio map-based path planning algorithm is proposed to
% obtain the flyable space that meets a certain signal to interference plus noise ratio (SINR) threshold by pre-constructing SINR map and the shortest flight.
Similarly, \cite{11031135} achieves continuous control of the UAV during
flight by constraining the communication rate to remain above the preset threshold.
\cite{11313542} calculates the signal strength and SINR based on the path loss model and integrates them into the path 
planning of multi-UAV networks.
Beyond static algorithms,  reinforcement learning (RL)-based and active sensing approaches have also been explored. 
For example, \cite{liu2019distributed} and \cite{8255824} use deep Q-networks and periodic environmental 
observations to jointly optimize the task 
completion rates and communication quality.
Furthermore, the UAV-assisted channel mapping is gaining significant attention.  
Notably, works such as \cite{11413848} and \cite{11143588} dynamically couple UAV trajectory planning with real-time environmental 
perceptions to achieve autonomous, high-precision three-dimensional spatial spectrum mapping. 
Despite these advancements, existing methods struggle to balance the environmental uncertainty and task complexity. 
These methods typically rely on perfectly known channel models as passive constraints, or 
focus exclusively on map reconstruction while ignoring the sequential constraints of practical physical tasks.
% However, these methods share two fundamental limitations. 
% Firstly, proactive optimization of communication quality is lacking, because it is typically treated as a passive 
% cost penalty or threshold constraint.
% Besides,  the sparse and incomplete nature of real-world wireless measurements is ignored under the assumption of 
% perfectly known channel models.

To this end, we consider introducing the channel knowledge map (CKM) to characterize the spatial 
distribution of received signal strength (RSS) during the inspection task.
% the wireless communication environment. 
% The CKM utilizes received signal strength (RSS) and other relevant indicators, such as the angle of arrival 
% and departure, to describe the spatial distribution of channel quality. 
% In mmWave or multiple-input multiple-output systems, the CKM facilitates the environment sensing 
% and beam alignment with low training overhead, effectively improving the communication efficiency \cite{9814544}. %9473871,
%Despite its potential
However, constructing an accurate CKM in urban environments remains challenging due to the scarcity 
of spatio-temporal channel measurement data, which hinders the practical communication-aware planning \cite{zeng2024tutorial,xu2024much,zhang2025strategic}. 
Therefore, we construct the diffusion-enhanced time-accumulated CKM, which yields a dense and stable channel representation without 
relying on extensive real-time sensing infrastructure by aggregating RSS measurements over time.
Building upon the time-accumulated CKM, a communication-aware UAV path planning strategy, 
termed as Graph Attention Network Soft Actor-Critic (GATSAC), is proposed to proactively avoid poor signal zones during urban inspection tasks.
% we propose a communication-aware UAV path planning strategy, 
% termed as Graph Attention Network-Soft Actor-Critic (GATSAC), to proactively avoid poor signal zones during urban inspection tasks.
% The approach ensures that the UAV trajectories are not only kinematically smooth but 
% also rigorously optimized for urban radio environments.
In detail, we reformulate the multi-target urban inspection issue as a traveling salesman problem (TSP) guided by graph attention network (GAT). 
The GAT effectively fuses spatial node features with CKM channel indicators, learning 
both local and global graph 
representations to infer an optimal visiting order that balances the flight duration and signal quality. 
Subsequently, a soft actor-critic (SAC)-based trajectory planner is designed to ensure the smooth and communication-efficient 
UAV trajectory based on the inferred sequence.

The rest of this paper is arranged as follows. 
Section \ref{section2} presents the system model and problem formulation.
In Section \ref{section3}, we design the CKM-driven path planning algorithm GATSAC. 
In Section \ref{section4}, we conduct simulations and analyze the results. 
Finally, Section \ref{section5} draws the conclusions.
\section{System Model and Problem Formulation}\label{section2}
\vspace{2mm}
In this section, the network model for the multi-UAV urban inspection scenario and the CKM model 
are presented,  followed by the problem formulation. 
\vspace{-0.5ex}
\subsection{Network Model}
\begin{figure}[t]
     \centering
     \includegraphics[width=0.9\linewidth]{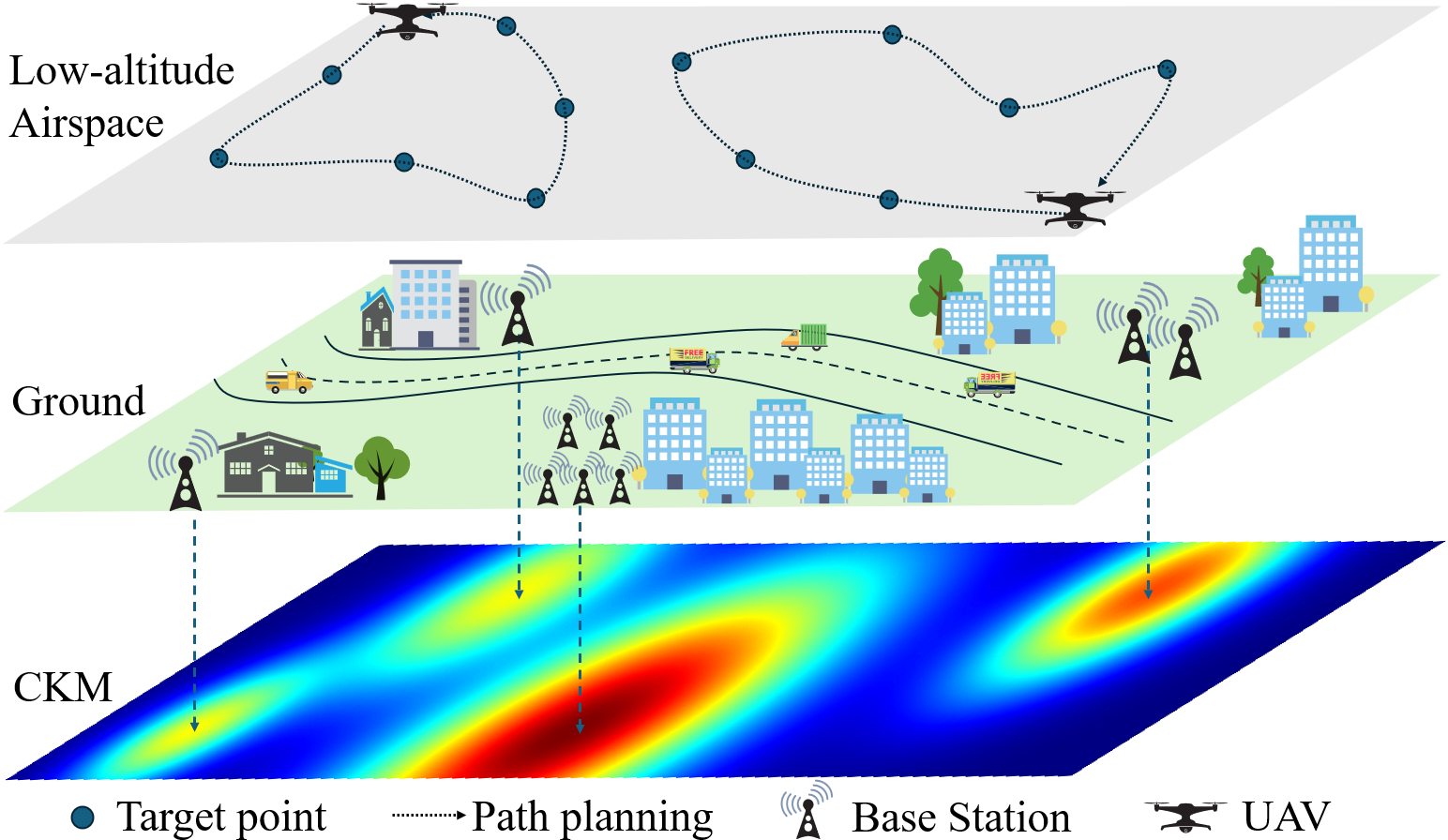}
     \captionsetup{font=small} 
     \caption{ Multi-UAV urban inspection scenario. The trajectory is generated based on 
     the constructed CKM.}
     \label{fig:scene}
\end{figure}

As shown in Fig. \ref{fig:scene}, we consider an urban UAV inspection scenario, where $M$ UAVs must visit $N$ target nodes 
distributed across an operational area $\mathcal{A}\subset\mathbb{R}^2$ at a fixed 
altitude $h$. 
The UAVs and target nodes are denoted as $\mathcal{U}=\{u_1,...,u_M\}$ and $\mathcal{V}=\{v_1,...,v_N\}$, respectively.
$u_i$ represents the $i$-th UAV, and $v_j$ represents the $j$-th target node. 
All UAVs are required to depart from and return to a depot $v_0$. 
% Additionally, each target node must be visited by at least one UAV. 
% The environment features dense buildings and complex propagation conditions that significantly 
% affect air-to-ground communication quality.

To enable the communication-aware path planning, we construct the CKM to capture the RSS spatial
distribution across $\mathcal{A}$. 
The CKM is leveraged to guide UAV trajectories toward regions with favorable channel conditions while minimizing the flight distance.
% Considering that the RSS is mainly determined by large-scale path loss, the entire environment can be regarded as 
% quasi-static, representing a typical urban area with buildings and obstacles that affect radio propagation. 
% Therefore, the CKM during the flight process can be regarded as hardly changing. 
Considering the coordination complexity, the task nodes are clustered into $M$ distinct groups based on the initial positions 
of UAVs, establishing a one-to-one correspondence which is expressed as 
\begin{equation}\mathbf{\mathcal{V}}=\bigcup_{m=1}^M\mathbf{\mathcal{V}_m}, \quad\mathbf{\mathcal{V}_m}\cap\mathbf{\mathcal{V}_n}=\emptyset,m\neq n.\end{equation}
$\mathbf{\mathcal{V}_m}$ represents the target node cluster with $N_m$ waypoints of the $m$-th UAV.
Each UAV independently serves one cluster, which transforms the multi-UAV mission into multiple single-UAV routing problems. 
Without loss of generality, we focus on one representative UAV $u_m$ with the cluster $\mathbf{\mathcal{V}_m}$. 
\vspace{-0.5ex}
\subsection{CKM Model}
% To characterize spatial channel variations, we define a CKM over the operational area $\mathcal{A}$. 
% The CKM is discretized into a uniform $H\times W$ grid as follows:
Considering that RSS is mainly determined by the large-scale path loss, we assume that the CKM % of the given area 
during flight can be regarded as quasi-static.
To facilitate spatial representation, the area $\mathcal{A}$ is discretized into a uniform grid of size $H\times W$, 
where $H$ and $W$ are determined by the physical dimensions and grid resolution. 
The mapping from continuous coordinates $(x,y)$ to discrete indices is denoted as $(g_x(x),g_y(y))$, given by
% To enable efficient spatial representation, we discretize $\mathcal{A}$ into a uniform grid of size $H\times W$. 
% $H$ and $W$ are determined by the physical dimensions of the area and the chosen grid resolution.
% We use $(g_x(x),g_y(y))$ to denote the discrete grid indices corresponding to the continuous physical coordinates $(x,y)$, detailed as 
\begin{equation}\label{Grid normalization1}g_x(x)=\mathrm{clip}\left(\left\lfloor\frac{x-X_{\min}}{X_{\max}-X_{\min}}\cdot(H-1)\right\rfloor,0,H-1\right),\end{equation}
% \newpage
and 
% \enlargethispage{-\baselineskip}
% \enlargethispage{-\baselineskip}
\begin{equation}\label{Grid normalization2}g_y(y)=\mathrm{clip}\left(\left\lfloor\frac{y-Y_{\min}}{Y_{\max}-Y_{\min}}\cdot(W-1)\right\rfloor,0,W-1\right),\end{equation} 
where $X_{\min}$, $X_{\max}$, $Y_{\min}$ and $Y_{\max}$ represent the spatial boundaries of the target area $\mathcal{A}$, 
and $\lfloor\cdot\rfloor$ indicates the floor operation.
The clip functions (\ref{Grid normalization1}) and (\ref{Grid normalization2}) restrict the resulting 
indices within the valid grid dimensions to prevent out-of-bound errors, 
and then each cell stores a channel quality indicator. 
% For the sake of convenience, $(g_x,g_y)$ is denoted as $(x,y)$ in the following text.
In addition, the RSS is used as the channel quality metric, i.e.,
\begin{equation}\mathcal{K}(g_x,g_y)\triangleq\text{RSS}(g_x,g_y).\end{equation} 
% Under the quasi-static assumption, accumulated RSS samples form a partially observed CKM used as ground truth for training, 
% enabling the model to learn the conditional distribution of the complete CKM from sparse measurements.
With the quasi-static assumption, RSS samples are accumulated over time to form a fully observed CKM, which serves as the complete CKM for training. 
Then, the model is trained to learn the conditional distribution of the complete CKM under sparse measurements, enabling the full CKM reconstruction.

% \vspace{-1.5ex}
 \subsection{Problem Formulation}
  To achieve efficient urban inspection with guaranteed communication quality,
  we formulate the trajectory optimization problem $\mathscr{P}0$ over the
  continuous trajectory $\mathbf{p}_m(t)$ of UAV $u_m$ with $N_m$ nodes in cluster $\mathcal{V}_m$.
For notational brevity, the spatial channel quality at the continuous UAV position is mapped to the 
discrete grid indices, which is defined as
\begin{equation}
\mathcal{K}(\mathbf{p}_m(t)) \triangleq \mathcal{K}\big(g_x(p_{x,m}(t)), \, g_y(p_{y,m}(t))\big),
\label{eq:channel_mapping}
\end{equation}
where $(p_{x,m}(t),p_{y,m}(t))$ is the instantaneous spatial coordinate of the UAV $u_m$ at time $t$.
%   For notational brevity, we define $\mathcal{K}(\mathbf{p}_m(t))\triangleq\mathcal{K}(g_x(p_{x,m}(t)), g_y(p_{y,m}(t)))$, where
%  $(p_{x,m}(t),p_{y,m}(t))$ is the instantaneous 2D spatial coordinate of the UAV $u_m$ at time $t$.
  Besides, the UAV follows a second-order kinematic model during flight, 
  where $\dot{\mathbf{p}}_m(t) = \boldsymbol{\xi}_m(t)$ and $\dot{\boldsymbol{\xi}}_m(t) =\mathbf{a}_m(t)$, 
  with $(\dot{\ })$ denoting time derivative.
  $\boldsymbol{\xi}_m(t)$ and $\mathbf{a}_m(t)$ are the velocity and acceleration vectors of UAV $u_m$,
  respectively.
%   Moreover, we use $\pi_m=\{v_{m,0},v_{i_1},\ldots,v_{i_{N_m}},v_{m,0}\}$ to represent the node visitation order corresponding to the continuous path optimization $\mathbf{p}_m(t)$,
%  where ${v_{m,0}}$ and $v_{i_k}$ denote the depot and $k$-th visited node of the UAV $u_m$, respectively.
%  In addition, to avoid nested minimization in the objective, we introduce an auxiliary variable $q_m$ to represent the worst communication quality along the trajectory of UAV $u_m$.
 In addition, to avoid nested minimization in the objective function, an auxiliary variable $q_m$ is introduced to represent the worst communication quality along the trajectory.
 The optimization problem is formulated as
\begin{subequations}\label{continuous problem formulation}
  \begin{align}
  \mathscr{P}0:\quad
  &\min_{\mathbf{p}_m(t),\,q_m}\;\int_0^{T_m}\|\boldsymbol{\xi}_m(t)\|\,dt - \lambda q_m\\
  \text{s.t.}\quad
    &\|\mathbf{p}_m(t_k)-\mathbf{p}_{v_k}\|\leq r,\quad k=1,\ldots,N_m,\label{constraint1}\\
    &\mathcal{K}(\mathbf{p}_m(t))\geq q_m,\quad\forall t\in[0,T_m],\label{constraintq}\\
    &\|\boldsymbol{\xi}_m(t)\|\leq\xi_{\max},\quad\forall t\in[0,T_m],\label{constraint3}\\
    &\|\mathbf{a}_m(t)\|\leq a_{\max},\quad\forall t\in[0,T_m],\label{constraint4}\\
    &\boldsymbol{\xi}_m(0)=\boldsymbol{\xi}_m(T_m)=\mathbf{0},\label{constraint5}\\
    &\mathbf{p}_m(0)=\mathbf{p}_m(T_m)=\mathbf{p}_{v_{m,0}},\label{constraint6}\\
    &\mathcal{K}_{\mathrm{min}} \leq q_m \leq \mathcal{K}_{\mathrm{max}},\label{constraint8}\\
    &0 \leq t_k \leq T_m,\quad k=1,\ldots,N_m,\label{constraint2}\\
    &\mathbf{p}_m(t)\in\mathcal{A},\quad\forall t\in[0,T_m].\label{constraint7}
  \end{align}
  \end{subequations}
  The objective function minimizes the total flight path length
  $\int_0^{T_m}\|\boldsymbol{\xi}_m(t)\|dt$ while maximizing the communication quality $q_m$,
  where $T_m$ represents the total task time, and $\lambda>0$ balances the flight efficiency and communication reliability.
  Constraint~(\ref{constraint1}) ensures that the UAV passes within capture
  radius $r$ of each waypoint $\mathbf{p}_{v_k}$ at time $t_k$. 
  In detail, $\mathbf{p}_{v_k}$ is the position of the waypoint $v_k$ in cluster $\mathcal{V}_m$, 
  and $t_k$ is the arrival time in continuous trajectory $\mathbf{p}_m(t)$ at $v_{k}$. 
  Constraint~(\ref{constraintq}) requires that the RSS value $\mathcal{K}(\mathbf{p}_m(t))$ 
  along the entire trajectory remains no less than $q_m$, for all times $t\in [0, T_m]$, 
  thereby imposing $q_m$ as a global lower bound on the communication quality throughout the flight. 
  Constraints~(\ref{constraint3}) and~(\ref{constraint4}) impose speed and
  acceleration limits, bounding the first and second derivatives of $\mathbf{p}_m(t)$, respectively. 
  Constraint~(\ref{constraint5}) enforces zero velocity at departure and return,
  and constraint~(\ref{constraint6}) requires the UAV to return to the depot $\mathbf{p}_{v_{m,0}}$.
  In constraint~(\ref{constraint8}), $q_m$ is restricted within the interval $[\mathcal{K}_{\mathrm{min}}, \mathcal{K}_{\mathrm{max}}]$,
  where $\mathcal{K}_{\mathrm{min}}$ and $\mathcal{K}_{\mathrm{max}}$ represent the minimum and maximum RSS values stored in the CKM, respectively.
  Constraint~(\ref{constraint2}) enforces that the arrival time of each waypoint is within the horizon $[0,T_m]$. 
  Constraint~(\ref{constraint7}) confines the trajectory within the operational area $\mathcal{A}$. 
%   Constraint~(\ref{constraint7}) requires $\pi$ to be a valid Hamiltonian cycle over all assigned nodes.

  $\mathscr{P}0$ is challenging to solve due to the non-differentiable CKM, nonconvex coverage constraints, and continuous kinematic constraints. Accordingly, we propose the CKM-driven GATSAC algorithm, which decomposes the
  problem into a discrete waypoint ordering stage and a continuous trajectory optimization stage solved in sequence.
\section{Algorithm Design}\label{section3}
\begin{figure}[t]
     \centering
     \includegraphics[width=0.9\linewidth]{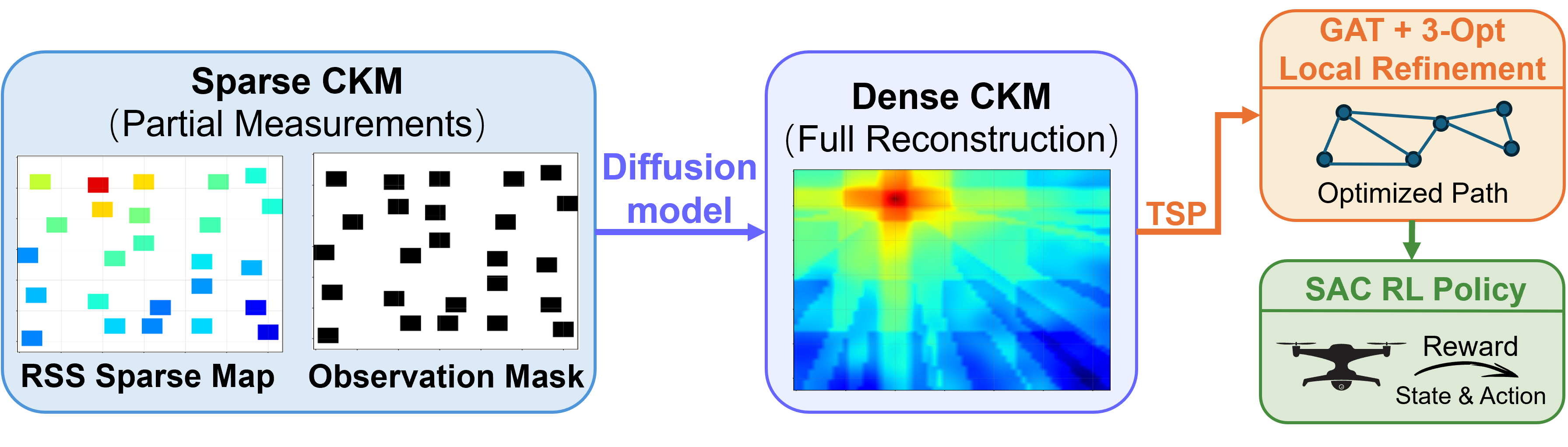}
     \captionsetup{font=small}
     \caption{\raggedright The overall framework of CKM-driven GATSAC algorithm.}
     \label{fig:framework}
\end{figure}

\vspace{2mm}
In this section, we present the CKM-driven GATSAC framework for communication-aware urban inspection, 
as illustrated in Fig. \ref{fig:framework}. 
We firstly present the construction of a complete CKM via a diffusion-based generative model. 
Then, we divide task-level path planning problem into two layers, which jointly addresses the 
communication-aware node ordering and continuous trajectory planning. 
In detail, a graph-based model firstly serves as the upper layer to determine the communication-aware visiting sequence of task nodes. 
Then, a continuous trajectory planner generates feasible UAV paths between consecutive nodes while avoiding regions with poor 
channel quality. 
The overall framework enables efficient and communication-aware urban UAV inspection tasks, detailed in Algorithm \ref{alg2}. 
\subsection{Time-Accumulated CKM Construction}
The CKM serves as the fundamental information layer that supports communication-aware trajectory planning.
\subsubsection{Temporal Accumulation under Quasi-Static Assumption}
  Exploiting the assumption of urban propagation quasi-static nature, RSS measurements
  across sequential episodes can be regarded as a sampling accumulation dimension and 
  aggregated to form progressively denser spatial maps.
  Speciﬁcally, we use $\delta_r^t(i,j)\triangleq\mathbb{I}\bigl[(i,j)=(g_x(t,r),\,g_y(t,r))\bigr]$
  to denote whether the channel measurements of receiver $r$ mapping to 
  grid cell $(i,j)$ at time step $t$ are available. 
  The accumulated signal sum and measurement count are updated via
  \begin{equation}
  \mathbf{S}_t(i,j)=\mathbf{S}_{t-1}(i,j)+\sum_{r\in\mathcal{R}_t}\mathrm{R}_{t,r}
  \cdot\delta_r^t(i,j),
  \end{equation}
  and 
  \begin{equation}
  \mathbf{C}_t(i,j)=\mathbf{C}_{t-1}(i,j)+\sum_{r\in\mathcal{R}_t}\delta_r^t(i,j),
  \end{equation}
  where $\mathbf{S}_t(i,j)$ and $\mathbf{C}_t(i,j)$ denote the accumulated signal
  sum and measurement count at grid $(i,j)$, $\mathcal{R}_t$ is the receiver set,
  and $\mathrm{R}_{t,r}$ is the RSS value of receiver $r$ at time step $t$.
% Exploiting the assumption of urban propagation quasi-static nature, RSS measurements collected across sequential episodes can be
%   regarded as a sampling accumulation dimension and aggregated to form progressively denser spatial maps. 
% Specifically, at each time step $t$, new observations are accumulated via
% % we adopt an incremental update mechanism for each time step and each grid position, which is detailed as
% % \begin{equation}
% % \begin{split}
% % \mathbf{S}_t(i,j)=\mathbf{S}_{t-1}(i,j)+\sum_{r\in\mathcal{R}_t}\mathrm{RSS}_{t,r}\cdot\mathbb{I}\left[(i,j) \\
% % =(g_x(t,r),g_y(t,r))\right], 
% % \end{split}
% % \end{equation}
% \begin{equation}
% \begin{split}
% \mathbf{S}_t(i,j)= &\mathbf{S}_{t-1}(i,j)+\\
% &\sum_{r\in\mathcal{R}_t}\mathrm{R}_{t,r}\cdot\mathbb{I}\Bigl[(i,j) = (g_x(t,r),g_y(t,r))\Bigr],
% \end{split}
% \end{equation}
% and 
% \begin{equation}
% \mathbf{C}_t(i,j)=\mathbf{C}_{t-1}(i,j)+\sum_{r\in\mathcal{R}_t}\mathbb{I}\Bigl[(i,j)=(g_x(t,r),g_y(t,r))\Bigr],
% \end{equation}
% where $\mathbf{S}_t(i,j)$ and $\mathbf{C}_t(i,j)$ denote the sum of accumulated signal strength and cumulative measurement count of grid $(i,j)$ at time step $t$, respectively.
% $\mathcal{R}_t$ is the receiver set at time step $t$, $\mathrm{R}_{t,r}$ indicates the RSS value of receiver $r$ 
% at time step $t$, 
% and $(g_x(t,r),g_y(t,r))$ denotes the grid position of the receiver $r$ at time step $t$. 
% $\mathbb{I}(\cdot)$ is an indicator function, which equals $1$ when channel measurements 
% of receiver $r$ at time step $t$ are available, and $0$ otherwise. 
An observation mask $\mathbf{M}_t(i,j)=\mathbb{I}\left[\mathbf{C}_t(i,j)>0\right]$ is defined to indicate the measured
  cells.
% In addition, we use masks to indicates which cells contain valid measurements, i.e.,
% \begin{equation}\mathbf{M}_t(i,j)=\mathbb{I}\left[\mathbf{C}_t(i,j)>0\right]\in\{0,1\}^{H\times W}.\end{equation}
Then, the accumulated CKM at time step $t$ can be expressed as 
\begin{equation}\mathbf{K}_t^\mathrm{raw}(i,j)=
\begin{cases}
\frac{\mathbf{S}_t(i,j)}{\mathbf{C}_t(i,j)}, & \mathrm{if~}\mathbf{M}_t(i,j)=1, \\
0, & \text{otherwise} 
,\end{cases}\end{equation}
where $\mathbf{K}_t^\mathrm{raw}(i,j)$ depicts the average sparse measurement value of grid $(i,j)$ at time step $t$.
RSS samples collected over multiple time steps are projected onto a unified spatial grid, forming a sparse CKM as
\begin{equation}\mathbf{K}_{\mathrm{accum}}(i,j)=\mathbf{K}_t^\mathrm{raw}(i,j),\end{equation}
% \begin{equation}\mathbf{K}_{\mathrm{accum}}\in\mathbb{R}^{H\times W},\end{equation}
where the observed grid cells are masked as a matrix, expressed as $\mathbf{K}_{\mathrm{accum}}$, and 
$\mathbf{K}_{\mathrm{accum}}(i,j)$ represents the value of the grid $(i,j)$ in the accumulated CKM.
After sufficient accumulation, certain scenes achieve full coverage, 
yielding complete maps that serve as ground truth for model training.
\subsubsection{Diffusion-Based CKM Generation}
While the time accumulation densifies the CKM, practical missions must plan trajectories based on the sparse observations from
 limited flight time, which typically cover only a small portion of the operational area. 
To enable reliable planning
 from such sparse data, we train a diffusion model on historical dense CKMs to learn the spatial propagation patterns.

Given the time-accumulated CKM as ground truth, represented as $\mathbf{K}_0=\mathbf{K}_{\mathrm{accum}}$, we
simulate sparse observations by randomly sampling to form training pairs, 
denoted as $(\mathbf{K}_{\text{sparse}},\mathbf{M})$. 
$\mathbf{K}_{\text{sparse}}$ is the sparse observations, and $\mathbf{M}\in\{0,1\}^{H\times W}$ is the observation mask. 
The forward diffusion process gradually
adds Gaussian noise to $\mathbf{K}_0$ over $T$ steps, i.e.,
\begin{equation}\mathbf{K}_t=\sqrt{\bar{\alpha}_t}\mathbf{K}_0+\sqrt{1-\bar{\alpha}_t}\boldsymbol{\epsilon}, \boldsymbol{\epsilon}\sim\mathcal{N}(0,\mathbf{I}),\end{equation} 
where $\bar{\alpha}_t$ controls the noise level at diffusion step $t$. 
Then, a UNet-based denoising network $\epsilon_\theta$ learns to predict the noise component conditioned on sparse observations, which is detailed as
\begin{equation}\epsilon_\theta=f_\theta(\mathbf{K}_t, t, \mathbf{K}_{\text{sparse}}, \mathbf{M}).\end{equation}
During the inference, the reverse process iteratively denoises from $\mathbf{K}_T \sim \mathcal{N}(0,\mathbf{I})$ to $\mathbf{K}_0$.
The observed radio measurements are strictly preserved by replacing $\mathbf{K}_t$ at each denoising step as
\begin{equation}\mathbf{K}_t\leftarrow\mathbf{K}_t\odot(1-\mathbf{M}_t)+\mathbf{K}_{\text{sparse}}\odot\mathbf{M}_t,\end{equation}
where $\mathbf{M}_t$ denotes the mask signal which take the value of 1 in grid with sparse observation, and 0 otherwise.
The training objective minimizes the noise prediction error as 
\begin{equation}\mathcal{L} = \mathbb{E}_{t,{\mathbf{K}_0},\boldsymbol{\epsilon}}\left[|\boldsymbol{\epsilon}-
\epsilon_\theta(\mathbf{K}_t, t, \mathbf{K}_{\text{sparse}}, \mathbf{M})|^2\right].\end{equation}
\subsection{Graph-Based TSP Solver}
We employ the GAT to better aggregate the channel state and the spatial information of target nodes
and 3-opt\cite{lancia2020finding} to dynamically adjust the node order in the TSP problem. 
Firstly, we combine the information of nodes and edges to construct the graph.
For a single UAV assigned with task nodes set $\mathcal{V}$, we construct a complete graph as
\begin{equation}\mathcal{G}=(\mathcal{V}\cup\{v_0\},\mathcal{E}),\end{equation}
where $v_0$ is the initial position of the UAV, and $\mathcal{E}$ contains all pairwise edges.
Additionally, the feature vector of each node $v_i\in\mathcal{V}$ is defined as
$\mathbf{x}_i=[x_i,y_i,k_i]$, consisting of spatial coordinates $(x_i,y_i)$ and the corresponding RSS value $k_i$ 
extracted from the CKM.
The edge feature between nodes $v_i$ and $v_j$ consists of the Euclidean distance and the minimum channel quality
$k_{ij}$ along the straight-line path between the two nodes.
% The feature of edge $(i,j)$ between nodes $v_i$ and $v_j$ includes the Euclidean distance  
% and the average channel quality $k_{ij}$ along the straight-line path between nodes $v_i$ and $v_j$.
Then, GAT is employed to aggregate the node and edge information and learn expressive node embeddings, depicted as
\begin{equation}\mathbf{h}_i\label{GAT}=\mathrm{GAT}(\mathbf{x}_i,\{\mathbf{x}_j,\mathbf{e}_{ij}\}_{j\in\mathcal{N}(i)}),\end{equation}
where $\mathbf{x}_j$ and $\mathbf{e}_{ij}$ represent the nodes and edges in the graph structure, respectively. 
Through iterative message passing, GAT captures both geometric relationships and communication-aware interactions among task nodes.
Based on the learned node embeddings, a decoding module generates a visiting sequence $\pi_0$ that starts and ends at the depot $v_0$.
The decoding process is implemented using the attention-based selection strategy.
% The resulting sequence approximates the solution of a dynamic TSP whose edge costs depend on both distance and channel quality. 
% Compared to classical TSP solvers, the proposed method enables efficient adaptation to spatially varying channel conditions.
\begin{algorithm}[t]
  \caption{CKM-Driven GATSAC for UAV Urban Inspection}
  \label{alg2}
  \begin{algorithmic}[1]
  \REQUIRE Target nodes $\mathcal{V}=\{v_1,\ldots,v_N\}$, depot $v_0$,
           dense CKM $\mathbf{K}^{\text{CKM}}$, trained GAT $\phi^*$,
           and trained SAC policy $\pi_\theta^*$.
  \ENSURE  UAV trajectory $\mathbf{p}(t)$ for all UAVs.
  \STATE Partition $\mathcal{V}$ into $M$ clusters $\{\mathcal{V}_1,\ldots,\mathcal{V}_M\}$ of $N_m$ waypoints via K-means clustering.
  \FOR{each UAV $u_m$ with assigned cluster $\mathcal{V}_m$}
      \STATE Construct graph $\mathcal{G}_m=(\mathcal{V}_m\cup\{v_0\},\mathcal{E}_m)$.
      \STATE Extract node features $\mathbf{x}_i=[x_i,y_i,k_i]$ and edge features $[d_{ij},k_{ij}]$ from $\mathbf{K}^{\text{CKM}}$.
      \STATE Compute node embeddings according to (\ref{GAT}).% $\mathbf{h}_i=\mathrm{GAT}_{\phi^*}(\mathbf{x}_i,\{\mathbf{x}_j,\mathbf{e}_{ij}\}_{j\in\mathcal{N}(i)})$
      \STATE Decode initial visiting sequence $\pi_{m,0}$ via attention-based selection.
      \STATE Refine $\pi_{m,0}$ with 3-opt and obtain $\pi_m^*$.
      \STATE Convert $\pi_m^*$ to sequence $\{v_{m,0},v_{i_1},\ldots,v_{i_{N_m}},v_{m,0}\}$.
      \STATE Initialize UAV at $\mathbf{p}_m(0)=\mathbf{p}_{v_{m,0}}$, $\mathbf{\xi}_m(0)=\mathbf{0}$, and waypoint index $k\gets 1$.
      \WHILE{$k \leq N_m$}
          \STATE Observe state $s_t=[\tilde{p}_x,\tilde{p}_y,\tilde{\xi}_x,\tilde{\xi}_y,\tilde{g}_x,\tilde{g}_y,\tilde{d},q_{\text{norm}}]$.
          \STATE Inquire $\mathbf{K}^{\text{CKM}}$ at current position to obtain $q_{\text{norm}}$.
          \STATE Select action $\mathbf{a_t}\sim\pi_\theta^*(\cdot|s_t)$.
          \STATE Execute $a_t$, then update $\mathbf{\xi}_m(t+1)$ and $\mathbf{p}_m(t+1)$.
          \IF{$\|\mathbf{p}_m(t)-\mathbf{p}_{v_{i_k}}\|\leq r$}
              \STATE Advance to next waypoint, $k\gets k+1$.
          \ENDIF
      \ENDWHILE
      \STATE Record trajectory $\mathbf{p}_m(t)$.
  \ENDFOR
  \RETURN $\{\mathbf{p}_m(t)\}_{m=1}^M$.
  \end{algorithmic}
  \end{algorithm}
\subsection{SAC-Based Continuous Trajectory Planning}
To further account for UAV kinematic constraints and environmental dynamics, the TSP solution is used as an initial policy for 
a SAC agent. Specifically, given the visiting sequence $\pi=\{v_0,v_{i_1},\ldots,v_{i_n},v_0\}$, 
the UAV is required to generate a continuous trajectory connecting consecutive nodes. 
The trajectory planning objective is to minimize the mission completion time $T$ while maintaining the communication quality.
In this work, the problem is formulated as a Markov decision process, which mainly consists of the following three parts:
\subsubsection{State Space}
At time step $t$, the state $s_t$ encodes the UAV normalized
position $(\tilde{c}_x,\tilde{c}_y)=({c_x}/{W},{c_y}/{H})$, velocity $(\tilde{\xi}_x,\tilde{\xi}_y)=({\xi_x}/{\xi_{\max}},{\xi_y}/{\xi_{\max}})$, next waypoint position $(\tilde{g}_x,\tilde{g}_y)$, 
distance to goal $\tilde{d}$, and the current RSS quality $q_{\text{norm}}\in[0,1]$ extracted from the CKM.
$(c_x,c_y)$ and $({\xi_x},{\xi_y})$ represent the mapping grid position and the velocity of the UAV.
\subsubsection{Action Space}
The continuous action $\mathbf{\tilde{a}}_t=[\tilde{a}_x,\tilde{a}_y]^T\in[-1,1]^2$ is a two-dimensional
  vector representing normalized acceleration commands along the $x$
  and $y$ axes, respectively. 
  $\mathbf{\tilde{a}}_t$ maps to the physical UAV acceleration via $\mathbf{a}_{t}=\mathbf{\tilde{a}}_t a_{\max}$,
  where $a_{\max}$ is the maximum UAV acceleration.
\subsubsection{Reward Function}
The reward function is designed to encourage the progress toward the target node while penalizing communication bottlenecks and excessive control 
effort, which is modeled as 
\begin{equation}r_t=\lambda_1 \Delta d_t^{\text{prog}}-\lambda_2\max(0,q_{\min}^{t-1}-q_t)\mathbb{I}[q_t<q_{\min}^{t-1}]-\lambda_3\Vert a_t\Vert ,\end{equation}
where $\Delta d_t^{\text{prog}}$ denotes the progress increment towards the target,  $q_t$ denotes the communication quality from the CKM, 
and $\Vert a_t\Vert$ penalizes the excessive control actions.
\subsubsection{Objective Function}
The policy network $\pi_\theta(a_t|s_t)$ parameterizes a Gaussian distribution over actions conditioned on the current 
state. 
The SAC maximizes the entropy-regularized expected return, i.e.,
\begin{equation}
\max_\theta\ \mathbb{E}_{\pi_\theta}\left[\sum_{t=0}^T\gamma^t(r_t+\alpha\mathcal{H}(\pi_\theta(\cdot|s_t)))\right],
\end{equation}
where $\gamma\in(0,1)$ is the discount factor, $\alpha>0$ is the entropy coefficient, and $\mathcal{H}(\cdot)$ denotes entropy. 
The optimized objective via the soft policy iteration is
\begin{equation}
\mathcal{L}_{\text{SAC}}=\mathbb{E}_{s_t,a_t}\left[\alpha\log\pi_\theta(a_t|s_t)-Q_\phi(s_t,a_t)\right],
\end{equation}
where $Q_\phi(s_t,a_t)$ is the soft Q-function approximated by a critic network. The entropy term encourages 
exploration while the Q-value guides exploitation, enabling stable and efficient learning.
\section{Simulation Analysis}\label{section4}
\vspace{2mm}
\begin{figure}[t]
     \centering
     \includegraphics[width=0.9\linewidth]{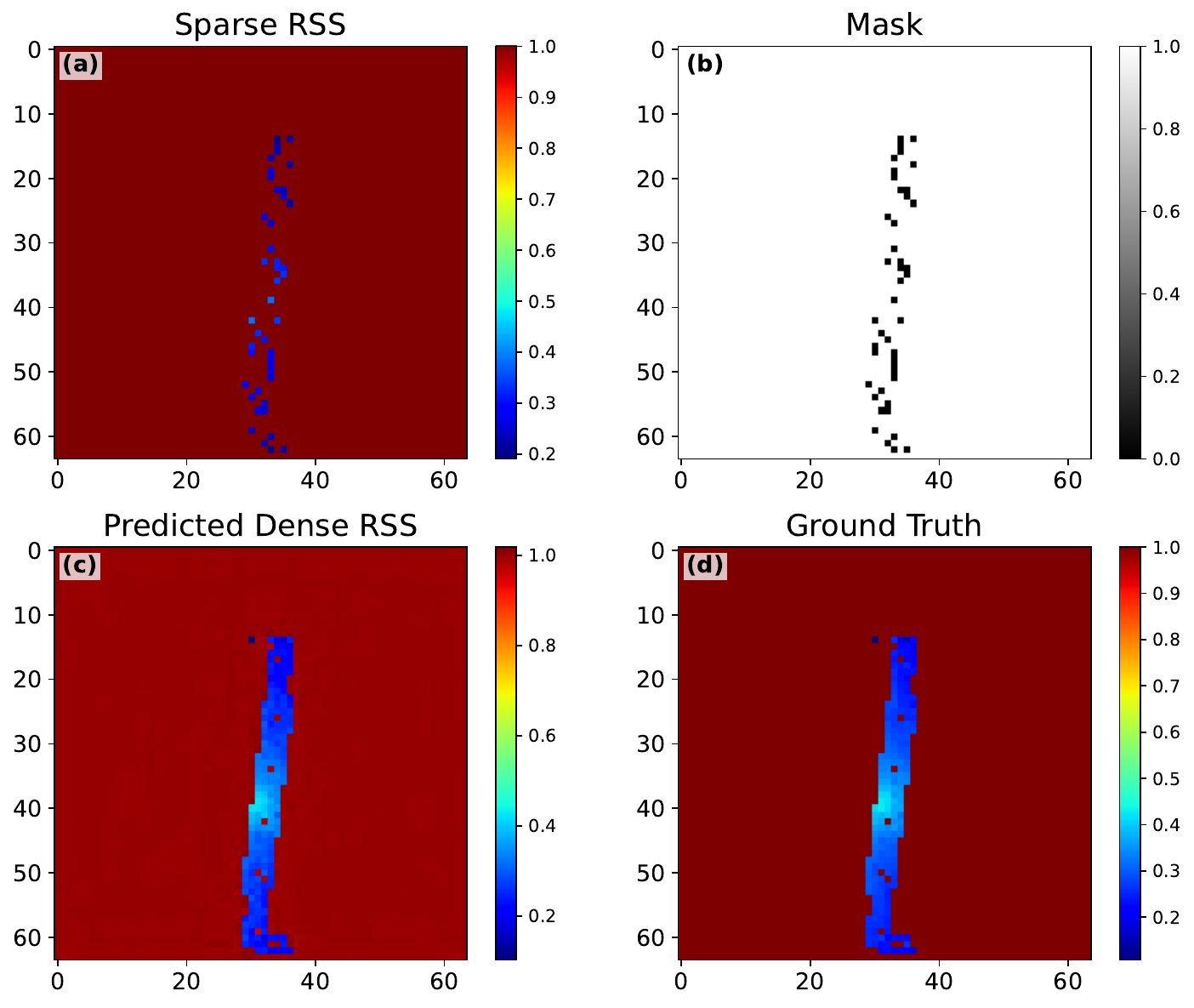}
     \captionsetup{font=small}
     \caption{\raggedright Diffusion-based time-accumulated CKM Construction.}
     \label{fig:CKM}
\end{figure}
% \begin{figure}[t]
%      \centering
%      \includegraphics[width=1\linewidth]{SAC Convergence.pdf}
%      \captionsetup{font=small}
%      \caption{\raggedright SAC Convergence.}
%      \label{fig:SAC}
% \end{figure}
\begin{figure}[t]
     \centering
     \includegraphics[width=0.9\linewidth]{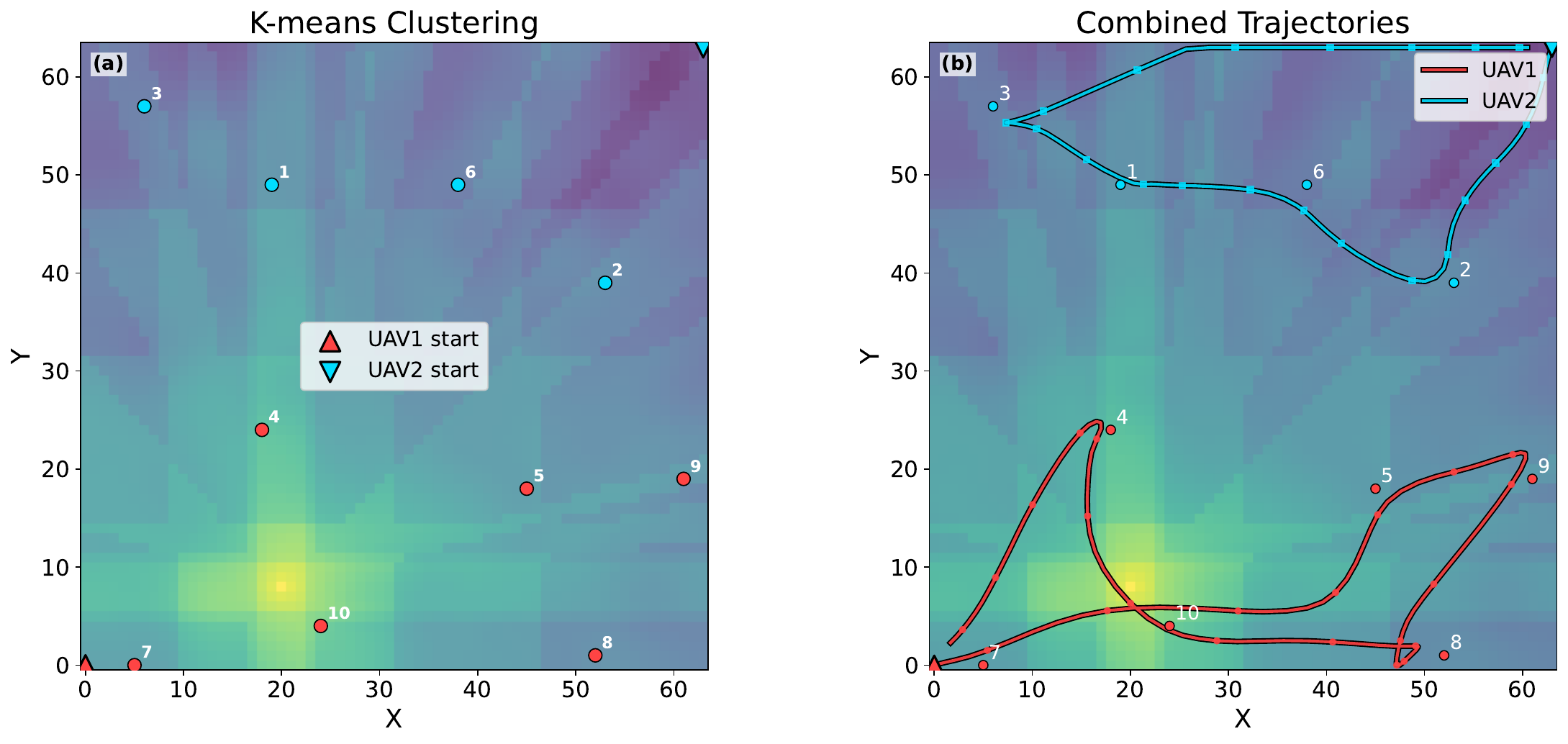}
     \captionsetup{font=small}
     \caption{\raggedright Multi-UAV cooperative  trajectory planning.}
     \label{fig:2UAV}
\end{figure}

\subsection{Experiment Setup}
To evaluate the effectiveness of the proposed CKM-driven GATSAC joint framework, comprehensive simulations are conducted in 
an urban UAV inspection scenario.
The simulation environment is constructed based on the Raymobtime urban dataset\cite{8503086}, which provides ray-tracing-based channel 
measurements and accurate geographical information. 
In detail, the target area for UAV network simulation is divided into $64\times 64$ grids.
% Due to sparse sampling and blockage effects, only a subset of grid points contains valid channel measurements, resulting in an 
% incomplete CKM.
Besides, to reduce the inference latency, we adopt the denoising diffusion implicit model sampling with $T=20$ steps, 
achieving a $50$-fold acceleration over the standard denoising diffusion probabilistic model while maintaining 
comparable reconstruction quality. 
To accurately evaluate the communication quality of a flight segment over a discretized CKM, the edge 
communication cost is defined as the minimum RSS among all grid cells intersected by the path between two nodes. 
% Moreover, a supercover line traversal is employed to ensure that all potentially obstructed cells are included, thereby preventing 
% underestimation of worst-case channel conditions. 
Aimed at demonstrating the performance of the proposed algorithm, five related methods are compared as follows.
\subsubsection{Distance-Only TSP without CKM}
This baseline solves the TSP using greedy and 3-opt algorithms based solely on Euclidean distance.%, entirely ignoring communication quality. 
Subsequently, its continuous trajectories are generated by a basic proportional controller. 
%The method solves the TSP using greedy and 3-opt methods based solely on Euclidean distance, ignoring communication quality, with trajectories generated by the proportional controller.
\subsubsection{Distance-Only TSP-SAC}%GAT-TSP without CKM
This method combines the distance-based TSP ordering with a pre-trained SAC agent for communication sensing and continuous trajectory planning. 
It enables the UAV to bypass poor communication areas, verifying the SAC module's independent contribution.
%Combines the distance-based TSP ordering with a pretrained SAC agent for communicaion sensing and continuous trajectory planning. 
%This enables the UAV to bypass poor communication areas, verifying the SAC module's independent contribution.
\subsubsection{TSP-A*}
After determining the node order, this baseline replaces our SAC-based continuous trajectory planner with the classical A* algorithm. 
It is implemented to evaluate the superiority of RL in continuous control.
%Replaces our SAC-based continuous trajectory planning with the classical A* algorithm after determining the node order.
\subsubsection{SAC without GAT-TSP Initialization}
Instead of utilizing GAT-TSP for sequence optimization, this approach adopts a random node visiting order and purely relies on the pre-trained SAC agent to achieve communication-aware continuous trajectory planning.
%Adopt random node sorting and use the pre-trained SAC agent to achieve communication-aware continuous trajectory planning.
\subsubsection{Joint SAC}
The method relies solely on the SAC agent to jointly optimize
  the discrete node ordering and continuous trajectory control,
  verifying the necessity of TSP initialization.
\subsection{Performance Analysis}
\begin{figure}[t]
     \centering
     \includegraphics[width=1\linewidth]{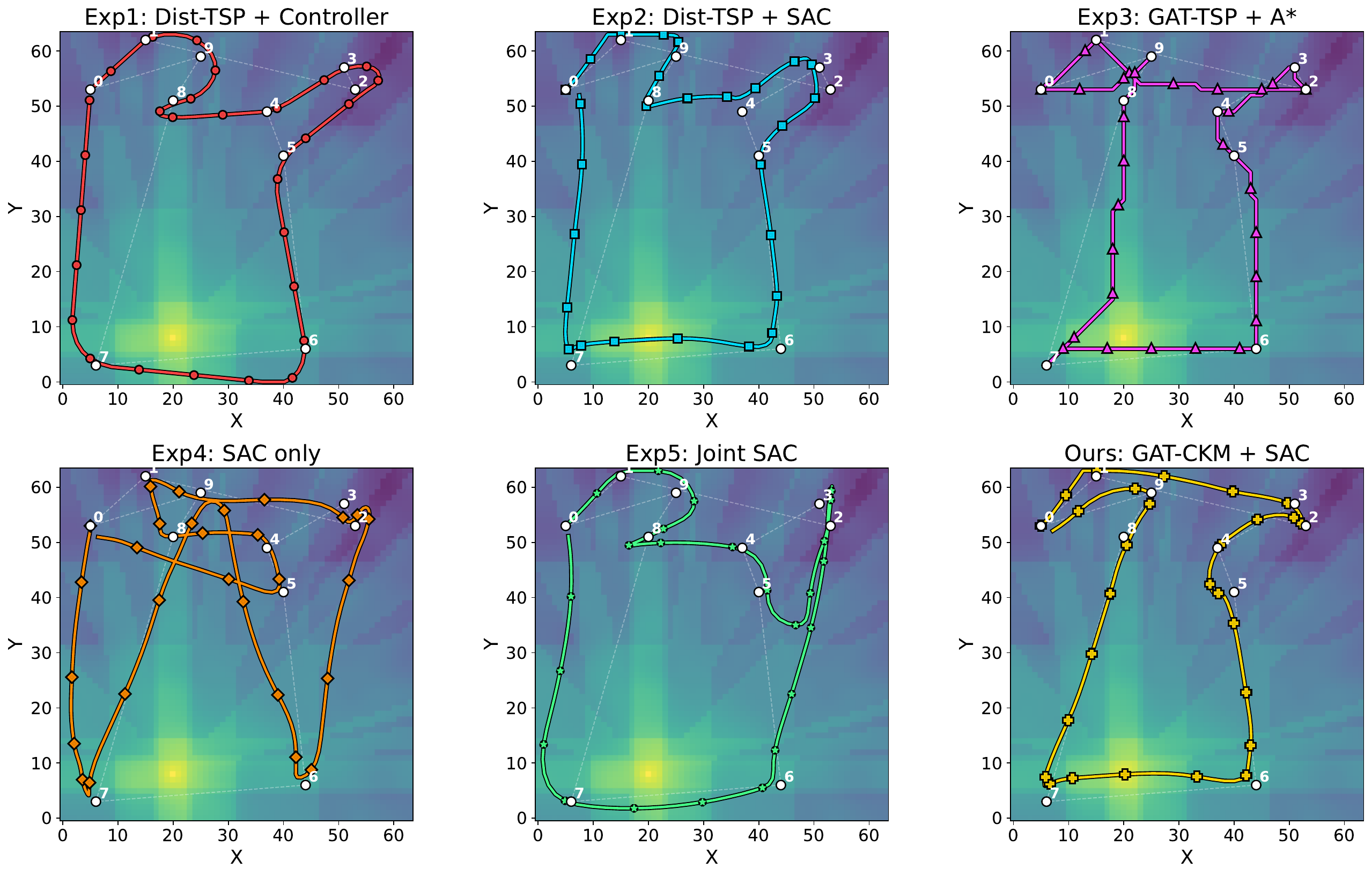}
     \captionsetup{font=small}
     \caption{\raggedright Trajectory comparisons of multiple planning methods.}
     \label{fig:comparison trajectory}
\end{figure}
\begin{figure}[t]
     \centering
     \includegraphics[width=1\linewidth]{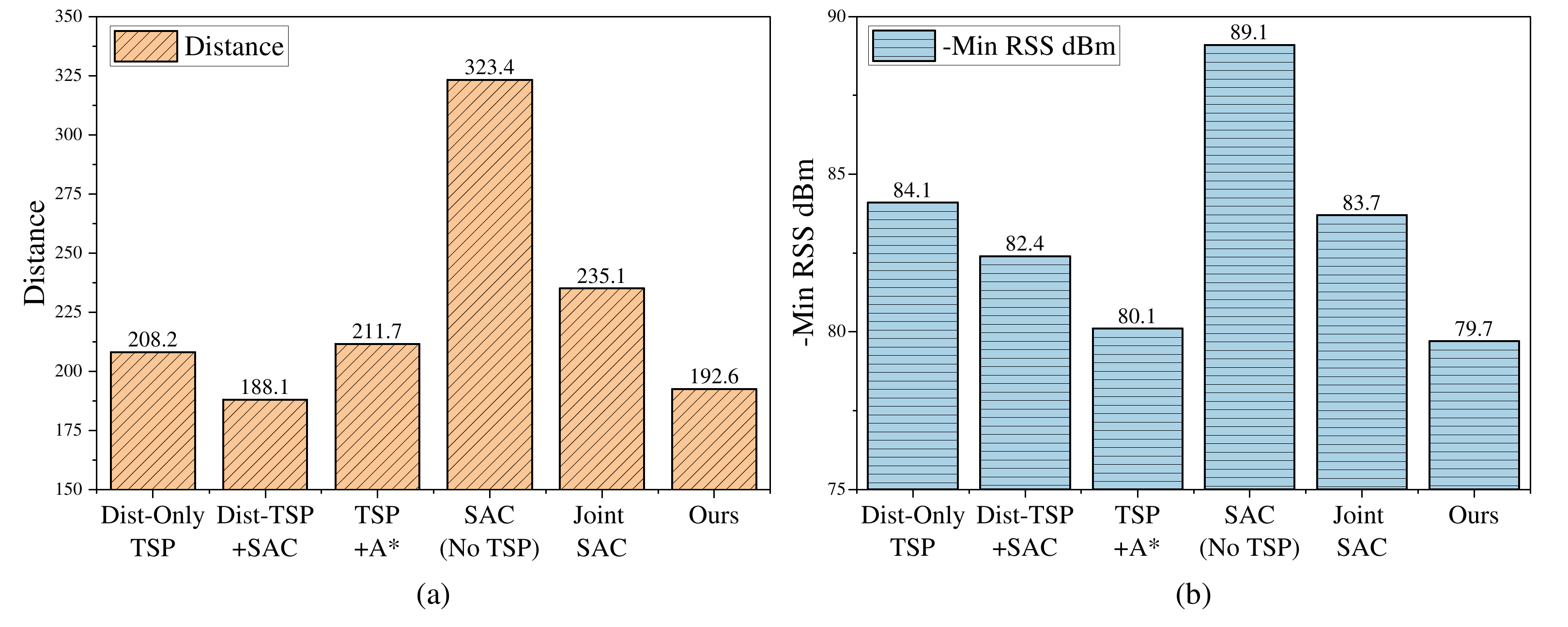}
     \captionsetup{font=small}
     \caption{\raggedright Comparison performance of six planning methods.}
     \label{fig:comparison}
\end{figure}
Fig.~\ref{fig:CKM} shows the process and performance of the CKM construction, where the discrete measurement points are shown in Fig.~\ref{fig:CKM}(a), 
and marked in the form of a mask in Fig.~\ref{fig:CKM}(b). 
Then, the reconstructed CKM in Fig.~\ref{fig:CKM}(c) is obtained through the pre-trained diffusion model, and Fig.~\ref{fig:CKM}(d) is the ground truth in the current region, which is relatively intuitive 
to observe that the error between the reconstructed CKM and ground truth is extremely small. 
Fig.~\ref{fig:2UAV} shows the results of trajectory planning for two UAVs, where Fig.~\ref{fig:2UAV}(a) shows the node division results 
for multiple UAVs, and Fig.~\ref{fig:2UAV}(b) is the final planning trajectory. 

Moreover, the trajectory planning results and performance comparison results of various algorithms are shown in 
Figs.~\ref{fig:comparison trajectory} and \ref{fig:comparison}. 
As observed in Fig.~\ref{fig:comparison trajectory}, compared with other algorithms, the planned trajectory of 
the proposed algorithm effectively avoids low-quality communication areas while maintaining a relatively low route cost. 
% In addition, Fig.\ref{fig:comparison} demonstrates that the proposed algorithm has the advantage of better balancing the 
% overall path length and communication performance, it has better performance for urban inspection tasks. 
Furthermore, it can be observed from Fig.~\ref{fig:comparison}(a) that the total flight distance of the proposed method is significantly shorter than 
most baseline methods. 
Besides, since the RSS in dBm is inherently negative, $-\text{MinRSS}$ is used as the vertical axis, 
such that a smaller value indicates better communication quality.
It is obvious that our algorithm achieves the optimal communication performance from Fig.~\ref{fig:comparison}(b). 
Compared with the Distance-Only TSP and TSP+A*, the CKM-driven GATSAC algorithm reduces the path length while incorporating communication awareness.
Additionally, although Dist-TSP+SAC yields the shortest distance, it does not explicitly optimize the communication 
quality and tends to traverse regions with poor communication quality due to the lack of channel awareness.
Meanwhile, the Joint SAC algorithm improves performance by incorporating learning-based planning, but it is still 
inferior to our proposed method due to the lack of structured global guidance. 
The SAC (No TSP) method performs the worst, which further highlights the importance of combining global planning with local optimization.
\section{Conclusions}\label{section5} 
\vspace{2mm}
In this paper, we propose the CKM-driven GATSAC algorithm for urban UAV inspection, simultaneously optimizing the flight efficiency and communication quality.
By constructing the diffusion-based time-accumulated CKM and integrating the graph-based TSP planning with SAC trajectory optimization, 
the method can effectively avoid low-quality communication regions while maintaining a relatively short flight distance. 
Simulation results validate the effectiveness of the proposed diffusion-based CKM and the hierarchical planning 
framework.
% \vspace{-2.7ex}
\vspace{2mm}
\bibliographystyle{IEEEtran}
\bibliography{REFERENCES.bib}%REFERENCES是bib文件的名字，里面放参考文献

\end{document}